\documentclass[english,american,conference, balance]{IEEEtran}
\usepackage[latin9]{inputenc}
\usepackage{babel}
\usepackage{array}
\usepackage{verbatim}
\usepackage{booktabs}
\usepackage{url}
\usepackage{graphicx}
\usepackage[unicode=true]
 {hyperref}

\makeatletter

\providecommand{\tabularnewline}{\\}

\usepackage{siunitx}
\usepackage{layouts}
\usepackage[table]{xcolor}
\usepackage{amsmath}
\definecolor{lightgray}{gray}{0.92}
\usepackage{graphicx}
\usepackage{listings}
\usepackage{xcolor}
\usepackage{url}
\usepackage{pifont}
\usepackage{multirow,bigdelim}
\usepackage{etoolbox}

\robustify\bfseries

\newcommand*\OK{\ding{51}}

\definecolor{codestrings}{rgb}{0.164,0,1}
\definecolor{codecomment}{rgb}{0.25,0.49,0.37}
\definecolor{codekeywords}{rgb}{0.8,0,0.33}
\definecolor{codebackground}{rgb}{0.95,0.95,0.95}
\lstdefinestyle{cblockstyle}{
inputencoding=utf8,
language=python,
extendedchars=true,
basicstyle=\ttfamily\footnotesize,
numbers=left,
  	numbersep=3pt,
framexleftmargin=2pt,
  	framerule=0pt,
  	frame=lines,
numberstyle=\tiny,
tabsize=2,
showstringspaces=false,
showspaces=false,
  keywordstyle=\bfseries\color{codekeywords},
  identifierstyle=\color{black},
  stringstyle=\color{codestrings},
  commentstyle=\color{codecomment},
  columns=fullflexible,
  abovecaptionskip=\medskipamount,
  belowcaptionskip=\medskipamount,
  backgroundcolor=\color{codebackground},
}
\lstdefinestyle{pblockstyle}{
inputencoding=utf8,
language=python,
extendedchars=true,
basicstyle=\ttfamily\footnotesize ,
numbers=left,
  	numbersep=3pt,
framexleftmargin=2pt,
  	framerule=0pt,
  	frame=lines,
numberstyle=\tiny,
tabsize=2,
showstringspaces=false,
showspaces=false,
  keywordstyle=\bfseries\color{codekeywords},
  identifierstyle=\color{black},
  stringstyle=\color{codestrings},
  commentstyle=\color{codecomment},
  columns=fullflexible,
  abovecaptionskip=\medskipamount,
  belowcaptionskip=\medskipamount,
  backgroundcolor=\color{codebackground},
}
\lstdefinestyle{clinestyle}{
tabsize=2,
frame=lines,
inputencoding=utf8,
language=C++,
keywordstyle=\bfseries\color{codekeywords},
identifierstyle=\color{black},
stringstyle=\color{codestrings},
commentstyle=\color{codecomment},
basicstyle=\ttfamily,
}
\lstnewenvironment{cblock}{\lstset{style=cblockstyle}}{}
\lstnewenvironment{clinee}{\lstset{style=clinestyle}}{}
\lstnewenvironment{pblock}{\lstset{style=pblockstyle}}{}

\makeatother

\begin{document}

\title{ViZDoom: A Doom-based AI Research Platform for Visual Reinforcement
Learning}

\author{\IEEEauthorblockN{Micha\l{} Kempka, Marek Wydmuch, Grzegorz Runc,
Jakub Toczek \& Wojciech Ja\'{s}kowski }\IEEEauthorblockA{Institute
of Computing Science, Poznan University of Technology, Pozna\'{n},
Poland\\
wjaskowski@cs.put.poznan.pl}}
\maketitle
\begin{abstract}
The recent advances in deep neural networks have led to effective
vision-based reinforcement learning methods that have been employed
to obtain human-level controllers in Atari 2600 games from pixel data.
Atari 2600 games, however, do not resemble real-world tasks since
they involve non-realistic 2D environments and the third-person perspective.
Here, we propose a novel test-bed platform for reinforcement learning
research from raw visual information which employs the first-person
perspective in a semi-realistic 3D world. The software, called ViZDoom,
is based on the classical first-person shooter video game, Doom. It
allows developing bots that play the game using the screen buffer.
ViZDoom is lightweight, fast, and highly customizable via a convenient
mechanism of user scenarios. In the experimental part, we test the
environment by trying to learn bots for two scenarios: a basic move-and-shoot
task and a more complex maze-navigation problem. Using convolutional
deep neural networks with Q-learning and experience replay, for both
scenarios, we were able to train competent bots, which exhibit human-like
behaviors.%
{} The results confirm the utility of ViZDoom as an AI research platform
and imply that visual reinforcement learning in 3D realistic first-person
perspective environments is feasible.

\end{abstract}

\section{Introduction\label{sec:Introduction}}

Visual signals are one of the primary sources of information about
the surrounding environment for living and artificial beings. While
computers have already exceeded humans in terms of raw data processing,
they still do not match their ability to interact with and act in
complex, realistic 3D environments. Recent increase in computing power
(GPUs), and the advances in visual learning (i.e., machine learning
from visual information) have enabled a significant progress in this
area. This was possible thanks to the renaissance of neural networks,
and deep architectures in particular. Deep learning has been applied
to many supervised machine learning tasks and performed spectacularly
well especially in the field of image classification \cite{NIPS2012_4824}.
Recently, deep architectures have also been successfully employed
in the reinforcement learning domain to train human-level agents to
play a set of Atari 2600 games from raw pixel information \cite{mnih-dqn-2015}.

Thanks to high recognizability and an easy-to-use software toolkit,
Atari 2600 games have been widely adopted as a benchmark for visual
learning algorithms. Atari 2600 games have, however, several drawbacks
from the AI research perspective. First, they involve only 2D environments.
Second, the environments hardly resemble the world we live in. Third,
they are third-person perspective games, which does not match a real-world
mobile-robot scenario. Last but not least, although, for some Atari
2600 games, human players are still ahead of bots trained from scratch,
the best deep reinforcement learning algorithms are already ahead
on average. Therefore, there is a need for more challenging reinforcement
learning problems involving first-person-perspective and realistic
3D worlds.

In this paper, we propose a software platform, ViZDoom\footnote{\url{http://vizdoom.cs.put.edu.pl}},
for the machine (reinforcement) learning research from raw visual
information. The environment is based on Doom, the famous first-person
shooter (FPS) video game. It allows developing bots that play Doom
using only the screen buffer. The environment involves a 3D world
that is significantly more real-world-like than Atari 2600 games.
It also provides a relatively realistic physics model. An agent (bot)
in ViZDoom has to effectively perceive, interpret, and learn the 3D
world in order to make tactical and strategic decisions where to go
and how to act. The strength of the environment as an AI research
platform also lies in its customization capabilities. The platform
makes it easy to define custom scenarios which differ by maps, environment
elements, non-player characters, rewards, goals, and actions available
to the agent. It is also lightweight \textendash{} on modern computers,
one can play the game at nearly $7000$ frames per second (the real-time
in Doom involves $35$ frames per second)  using a single CPU core,
which is of particular importance if learning is involved.

In order to demonstrate the usability of the platform, we perform
two ViZDoom experiments with deep Q-learning \cite{mnih-dqn-2015}.
The first one involves a somewhat limited 2D-like environment, for
which we try to find out the optimal rate at which agents should make
decisions. In the second experiment, the agent has to navigate a 3D
maze collecting some object and omitting the others. The results of
the experiments indicate that deep reinforcement learning is capable
of tackling  first-person perspective 3D environments\footnote{Precisely speaking, Doom is pseudo-3D or 2.5D.}.

FPS games, especially the most popular ones such as Unreal Tournament
\cite{6314567,6922494}, Counter-Strike \cite{5035619} or Quake III
Arena \cite{el2007hybrid}, have already been used in AI research.
However, in these studies agents acted upon high-level information
like positions of walls, enemies, locations of items, etc., which
are usually inaccessible to human players. Supplying only raw visual
information might relieve researchers of the burden of providing AI
with high-level information and handcrafted features. We also hypothesize
that it could make the agents behave more believable \cite{karpov:believablebots12}.
So far, there has been no studies on reinforcement learning from visual
information obtained from FPS games.

To date, there have been no FPS-based environments that allow research
on agents relying exclusively on raw visual information. This could
be a serious factor impeding the progress of vision-based reinforcement
learning, since engaging in it requires a large amount of programming
work. Existence of a ready-to-use tool facilitates conducting experiments
and focusing on the goal of the research.

\section{Related Work\label{sec:Related-Work}}

One of the earliest works on visual-based reinforcement learning is
due to Asada et al. \cite{asada1994vision,asada1996purposive}, who
trained robots various elementary soccer-playing skills. Other works
in this area include teaching mobile robots with visual-based $Q$-learning
\cite{gaskett2000reinforcement}, learning policies with deep auto-encoders
and batch-mode algorithms \cite{lange2010deep}, neuroevolution for
a vision-based version of the mountain car problem \cite{cuccu2011intrinsically},
and compressed neuroevolution with recurrent neural networks for vision-based
car simulator \cite{koutnik2014evolving}. Recently, Mnih et al. have
shown a deep Q-learning method for learning Atari 2600 games from
visual input \cite{mnih-dqn-2015}.

Different first-person shooter (FPS) video games have already been
used either as AI research platforms, or application domains. The
first academic work on AI in FPS games is due to Geisler \cite{geisler2002empirical}.
It concerned modeling player behavior in Soldier of Fortune 2. Cole
used genetic algorithms to tune bots in Counter Strike \cite{cole2004using}.
Dawes \cite{dawes2005towards} identified Unreal Tournament 2004 as
a potential AI research test-bed. El Rhalib studied weapon selection
in Quake III Arena \cite{el2007hybrid}. %
Smith devised a RETALIATE reinforcement learning algorithm for optimizing
team tactics in Unreal Tournament \cite{smith2007retaliate}. SARSA($\lambda$),
another reinforcement learning method, was the subject of research
in FPS games \cite{5672586,6314567}. Recently, continuous and reinforcement
learning techniques were applied to learn the behavior of tanks in
the game BZFlag \cite{smith2014continuous}.

As far as we are aware, to date, there have been no studies that employed
the genre-classical Doom FPS. Also, no previous study used raw visual
information to develop bots in first-person perspective games with
a notable exception of the Abel's et al. work on Minecraft \cite{DBLP:journals/corr/AbelADKS16}.

\section{ViZDoom Research Platform\label{sec:Platform}}

\subsection{Why Doom?}

Creating yet another 3D first-person perspective environment from
scratch solely for research purposes would be somewhat wasteful \cite{trenholme2008computer}.
Due to the popularity of the first-person shooter genre, we have decided
to use an existing game engine as the base for our environment. We
concluded that it has to meet the following requirements:
\begin{enumerate}
\item based on popular open-source 3D FPS game (ability to modify the code
and the publication freedom),
\item lightweight (portability and the ability to run multiple instances
on a single machine),
\item fast (the game engine should not be the learning bottleneck),
\item total control over the game's processing (so that the game can wait
for the bot decisions or the agent can learn by observing a human
playing),
\item customizable resolution and rendering parameters,
\item multiplayer games capabilities (agent vs. agent and agent vs. human),
\item easy-to-use tools to create custom scenarios,
\item ability to bind different programming languages (preferably written
in C++),
\item multi-platform.
\end{enumerate}
\begin{figure}
\centering{}\includegraphics[width=0.9\columnwidth]{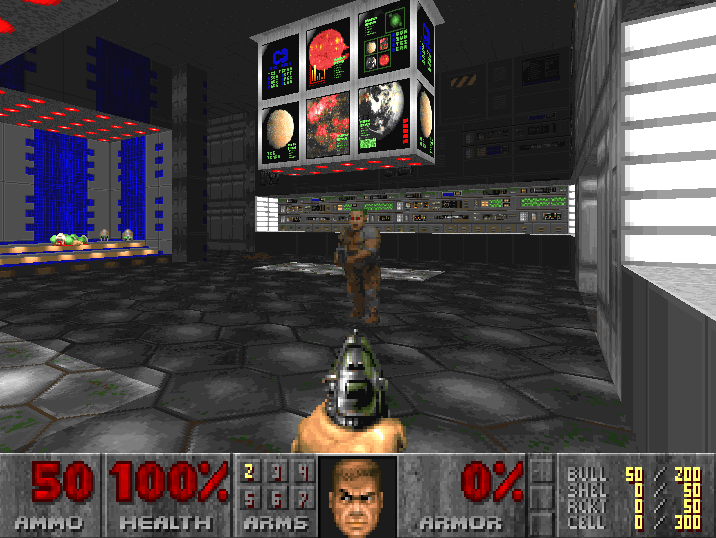}
\caption{\label{fig:doom}Doom's first-person perspective.}
\end{figure}

In order to make the decision according to the above-listed criteria,
we have analyzed seven recognizable FPS games: Quake III Arena, Doom
3, Half-Life 2, Unreal Tournament 2004, Unreal Tournament and Cube.
Their comparison is shown in Table~\ref{tab:engines}. Some of the
features listed in the table are objective (e.g., `scripting') and
others are subjective (``code complexity''). Brand recognition was
estimated as the number (in millions) of Google results (as of 26.04.2016)
for phrases ``game \textless{}gamename\textgreater{}'', where \textless{}gamename\textgreater{}
was `doom', `quake', `half-life', `unreal tournament' or `cube'. The
game was considered as low-resolution capable if it was possible to
set the resolution to values smaller than $640\times480$.

\begin{table*}
\caption{\label{tab:engines}Overview of 3D FPS game engines considered.}

\centering{}%
\begin{tabular}{lcccc>{\centering}p{1.5cm}>{\centering}p{1.5cm}c}
\toprule 
Features / Game & Doom & Doom 3  & Quake III: Arena  & Half-Life 2  & Unreal Tournament 2004   & Unreal Tournament   & Cube\tabularnewline
\midrule
Game Engine  & ZDoom\cite{zdoom-wiki} & id tech 4  & ioquake3  & Source  & Unreal Engine 2  & Unreal Engine 4  & Cube Engine \tabularnewline
Release year  & 1993  & 2003  & 1999  & 2004  & 2004  & not yet  & 2001 \tabularnewline
Open Source  & \OK  & \OK  & \OK  &  &  & \OK  & \OK \tabularnewline
License  & GPL  & GPLv3  & GPLv2  & Proprietary  & Proprietary & Custom  & ZLIB \tabularnewline
Language  & C++  & C++  & C  & C++  & C++  & C++  & C++ \tabularnewline
\midrule 
DirectX  &  & \OK  &  & \OK  &  & \OK  & \tabularnewline
OpenGL  & \OK\footnotemark  & \OK  & \OK  & \OK  & \OK  & \OK  & \OK \tabularnewline
Software Render  & \OK  &  &  &  &  &  & \tabularnewline
\midrule
Windows  & \OK  & \OK  & \OK  & \OK  & \OK  & \OK  & \OK \tabularnewline
Linux  & \OK  & \OK  & \OK  & \OK  & \OK  & \OK  & \OK \tabularnewline
Mac OS  & \OK  & \OK  & \OK  & \OK  & \OK  & \OK  & \tabularnewline
\midrule
Map editor  & \OK  & \OK  & \OK  & \OK  & \OK  & \OK  & \OK \tabularnewline
Screen buffer access  & \OK  & \OK  & \OK  &  &  & \OK  & \OK \tabularnewline
Scripting  & \OK  & \OK  &  & \OK  & \OK  & \OK  & \OK \tabularnewline
Multiplayer mode & \OK  & \OK  & \OK  &  & \OK  & \OK  & \OK \tabularnewline
Small resolution & \OK  & \OK  & \OK  & \OK  & \OK  & \OK  & \OK \tabularnewline
\midrule
Custom assets  & \OK  & \OK  & \OK & \OK  & \OK  & \OK  & \OK \tabularnewline
Free original assets  &  &  &  &  &  & \OK  & \OK \tabularnewline
\midrule
System requirements  & Low  & Medium  & Low  & Medium  & Medium  & High  & Low\tabularnewline
Disk space & 40MB  & 2GB  & 70MB  & 4,5GB  & 6GB  & \textgreater{}10GB  & 35MB \tabularnewline
Code complexity  & Medium & High  & Medium  & - & -  & High  & Low \tabularnewline
\midrule 
Active community  & \OK  & \OK  & \OK  & \OK  &  & \OK  & \tabularnewline
\midrule
Brand recognition & \multicolumn{2}{c}{31.5} & 16.8 & 18.7 & \multicolumn{2}{c}{1.0} & 0.1\tabularnewline
\bottomrule
\end{tabular}
\end{table*}

Some of the games had to be rejected right away in spite of high general
appeal. Unreal Tournament 2004 engine is only accessible by the Software
Development Kit and it lacks support for controlling the speed of
execution and direct screen buffer access. The game has not been prepared
to be heavily modified.

Similar problems are shared by Half-Life 2 despite the fact that the
Source engine is widely known for modding capabilities. It also lacks
direct multiplayer support. Although the Source engine itself offers
multiplayer support, it involves client-server architecture, which
makes synchronization and direct interaction with the engine problematic
(network communication).

\footnotetext{GZDoom, the ZDoom's fork, is OpenGL-based.}The client-server
architecture was also one the reasons for rejection of Quake III:
Arena. Quake III also does not offer any scripting capabilities, which
are essential to make a research environment versatile. The rejection
of Quake was a hard decision as it is a highly regarded and playable
game even nowadays but this could not outweigh the lack of scripting
support.

The latter problem does not concern Doom 3 but its high disk requirements
were considered as a drawback. Doom 3 had to be ignored also because
of its complexity, Windows-only tools, and OS-dependent rendering
mechanisms. Although its source code has been released, its community
is dispersed. As a result, there are several rarely updated versions
of its sources.

The community activity is also a problem in the case of Cube as its
last update was in August 2005. Nonetheless, the low complexity of
its code and the highly intuitive map editor would make it a great
choice if the engine was more popular.

Unreal Tournament, however popular, is not as recognizable as Doom
or Quake but it has been a primary research platform for FPS games
\cite{5586059,6046867}. It also has great capabilities. Despite its
active community and the availability of the source code, it was rejected
due to its high system requirements.

Doom (see Fig.~\ref{fig:doom}) met most of the requirements and
allowed to implement features that would be barely achievable in other
games, e.g., off-screen rendering and custom rewards. The game is
highly recognizable and runs on the three major operating systems.
It was also designed to work in $320\times240$ resolution and despite
the fact that modern implementations allow bigger resolutions, it
still utilizes low-resolution textures. Moreover, its source code
is easy-to-understand.

The unique feature of Doom is its software renderer. Because of that,
it could be run without the desktop environment (e.g., remotely in
a terminal) and accessing the screen buffer does not require transferring
it from the graphics card.

Technically, ViZDoom is based on the modernized, open-source version
of Doom's original engine \textemdash{} ZDoom, which is still actively
supported and developed.

\subsection{Application Programming Interface (API)\label{subsec:api}}

ViZDoom API is flexible and easy-to-use. It was designed with reinforcement
and apprenticeship learning in mind, and therefore, it provides full
control over the underlying Doom process. In particular, it allows
retrieving the game's screen buffer and make actions that correspond
to keyboard buttons (or their combinations) and mouse actions. Some
game state variables such as the player's health or ammunition are
available directly. 

ViZDoom's API was written in C++. The API offers a myriad of configuration
options such as control modes and rendering options. In addition to
the C++ support, bindings for Python and Java have been provided.
The Python API example is shown in Fig.~\ref{fig:PythonExample}.

\begin{figure}
\begin{pblock}  
from vizdoom import *
from random import choice 
from time import sleep, time

game = DoomGame() 
game.load_config("../config/basic.cfg")  
game.init()

# Sample actions. Entries correspond to buttons:
# MOVE_LEFT, MOVE_RIGHT, ATTACK 
actions = [[True, False, False],            
			[False, True, False], [False, False, True]]
# Loop over 10 episodes.
for i in range(10): 	
	game.new_episode() 	
	while not game.is_episode_finished():  		
		# Get the screen buffer and and game variables 		
		s = game.get_state() 		
		img = s.image_buffer 		
		misc = s.game_variables 
		# Perform a random action: 		
		action = choice(actions) 		
		reward = game.make_action(action)
		# Do something with the reward...
	
	print("total reward:", game.get_total_reward()) 	
\end{pblock}\caption{\label{fig:PythonExample}Python API example}
\end{figure}

\subsection{Features}

ViZDoom provides features that can be exploited in different kinds
of AI experiments. The main features include different control modes,
custom scenarios, access to the depth buffer and off-screen rendering
eliminating the need of using a graphical interface.

\subsubsection{Control modes}

ViZDoom implements four control modes: i) synchronous player, ii)
synchronous spectator, iii) asynchronous player, and iv) asynchronous
spectator. 

In asynchronous modes, the game runs at constant $35$ frames per
second and if the agent reacts too slowly, it can miss some frames.
Conversely, if it makes a decision too quickly, it is blocked until
the next frame arrives from the engine. Thus, for reinforcement learning
research, more useful are the synchronous modes, in which the game
engine waits for the decision maker. This way, the learning system
can learn at its pace, and it is not limited by any temporal constraints.

Importantly, for experimental reproducibility and debugging purposes,
the synchronous modes run deterministically.

In the player modes, it is the agent who makes actions during the
game. In contrast, in the spectator modes, a human player is in control,
and the agent only observes the player's actions.

In addition, ViZDoom provides an asynchronous multiplayer mode, which
allows games involving up to eight players (human or bots) over a
network.

\subsubsection{Scenarios}

One of the most important features of ViZDoom is the ability to run
custom scenarios. This includes creating appropriate maps, programming
the environment mechanics (``when and how things happen''), defining
terminal conditions (e.g., ``killing a certain monster'', ``getting
to a certain place'', ``died''), and rewards (e.g., for ``killing
a monster'', ``getting hurt'', ``picking up an object''). This
mechanism opens endless experimentation possibilities. In particular,
it allows creating a scenario of a difficulty which is on par with
the capabilities of the assessed learning algorithms.

Creation of scenarios is possible thanks to easy-to-use software tools
developed by the Doom community. The two recommended free tools include
Doom Builder 2 and SLADE 3. Both are visual editors, which allow
defining custom maps and coding the game mechanics in Action Code
Script. They also enable to conveniently test a scenario without leaving
the editor.

ViZDoom comes with a few predefined scenarios. Two of them are described
in Section \ref{sec:Experiment}.

\subsubsection{Depth Buffer Access}

\begin{figure}
\centering{}\includegraphics[width=0.9\columnwidth]{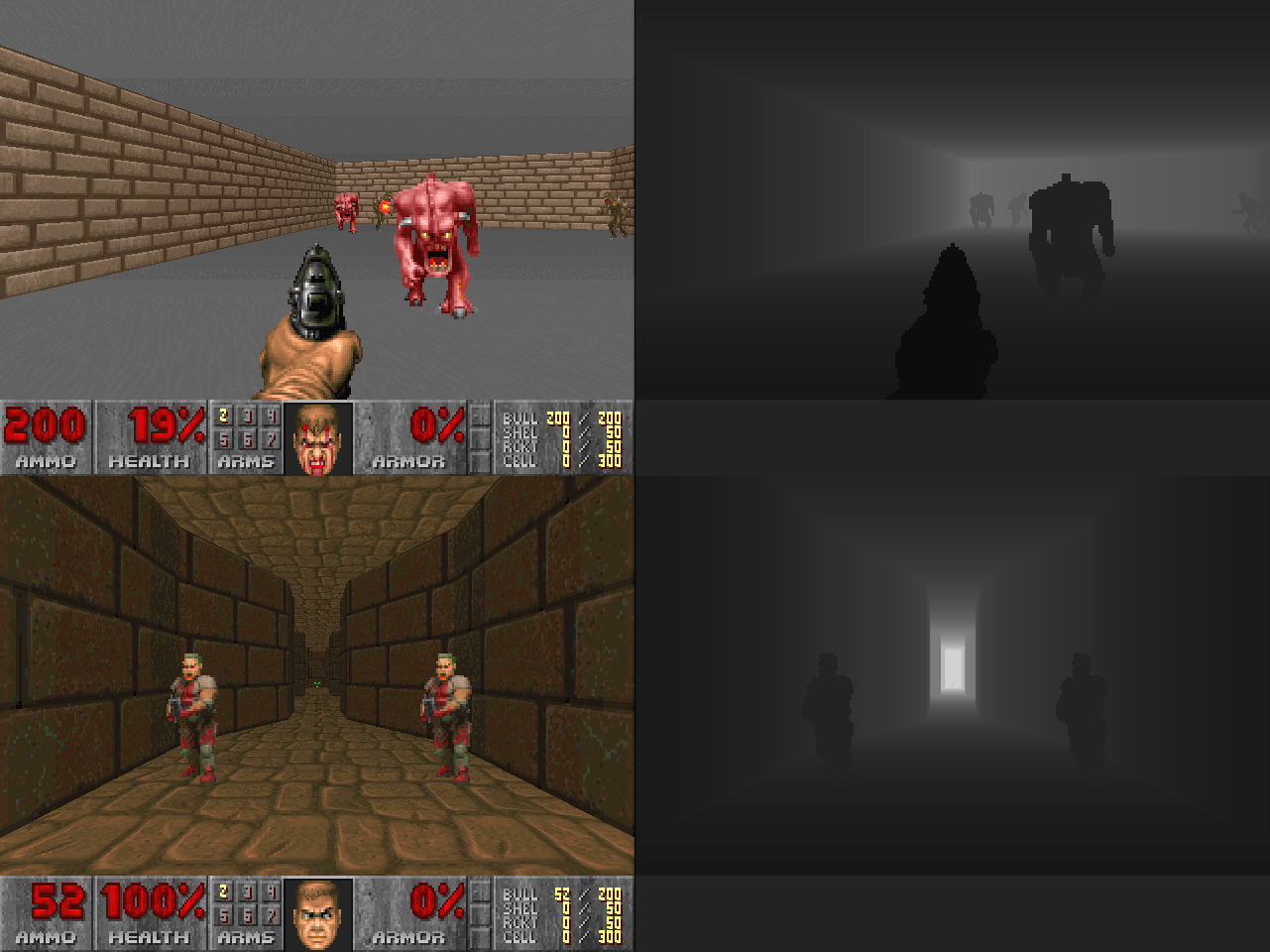}
\caption{\label{fig:zbuffer}ViZDoom allows depth buffer access.}
\end{figure}

ViZDoom provides access to the renderer's depth buffer (see Fig.~\ref{fig:zbuffer}),
which may help an agent to understand the received visual information.
This feature gives an opportunity to test whether the learning algorithms
can autonomously learn the whereabouts of the objects in the environment.
The depth information can also be used to simulate the distance sensors
common in mobile robots.

\subsubsection{Off-Screen Rendering and Frame Skipping\label{subsec:Off-Screen-Skipping}}

To facilitate computationally heavy machine learning experiments,
we equipped ViZDoom with off-screen rendering and frame skipping features.
Off-screen rendering lessens the performance burden of actually showing
the game on the screen and makes it possible to run the experiments
on the servers (no graphical interface needed). Frame skipping, on
the other hand, allows omitting rendering selected frames at all.
Intuitively, an effective bot does not have to see every single frame.
We explore this issue experimentally in Section \ref{sec:Experiment}.

\subsection{ViZDoom's Performance}

\begin{figure}
\centering{}\includegraphics[width=1\columnwidth]{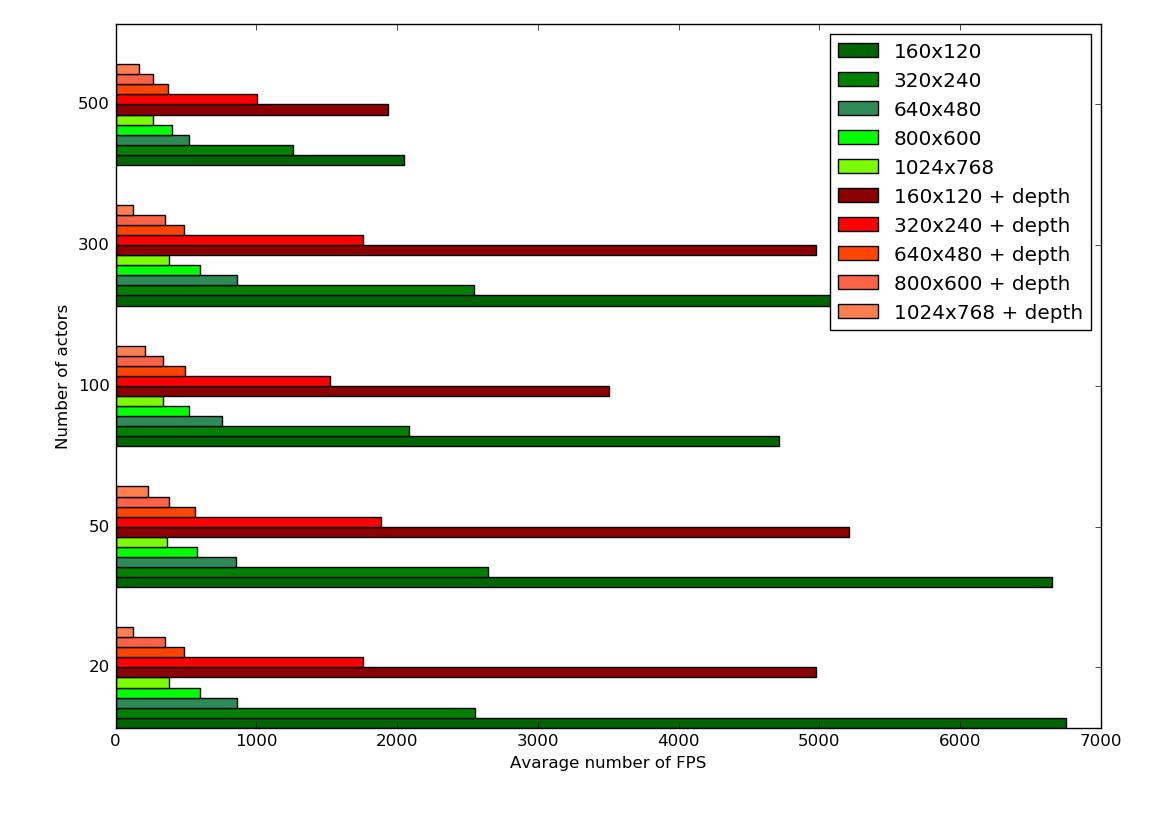}
\caption{\label{fig:fps_test}ViZDoom performance. ``depth'' means generating
also the depth buffer.}
\end{figure}

The main factors affecting ViZDoom performance are the number of the
actors (like items and bots), the rendering resolution, and computing
the depth buffer. Fig.~\ref{fig:fps_test} shows how the number of
frames per second depends on these factors. The tests have been made
in the synchronous player mode on Linux running on Intel Core i7-4790k.
ViZDoom uses only a single CPU core.

The performance test shows that ViZDoom can render nearly $7000$
low-resolution frames per second. The rendering resolution proves
to be the most important factor influencing the processing speed.
In the case of low resolutions, the time needed to render one frame
is negligible compared to the backpropagation time of any reasonably
complex neural network.

\section{Experiments\label{sec:Experiment}}

\subsection{Basic Experiment}

The primary purpose of the experiment was to show that reinforcement
learning from the visual input is feasible in ViZDoom. Additionally,
the experiment investigates how the number of skipped frames (see
Section \ref{subsec:Off-Screen-Skipping}) influences the learning
process.

\begin{figure}
\centering{}\includegraphics[width=0.9\columnwidth]{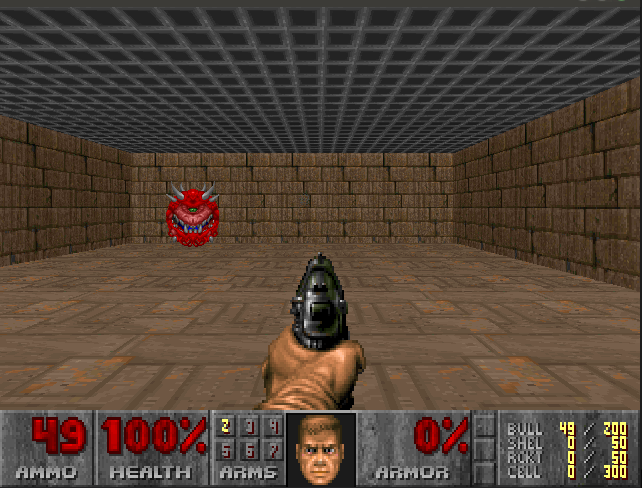}
\caption{\label{fig:basic}The basic scenario }
\end{figure}

\subsubsection{Scenario}

This simple scenario takes place in a rectangular chamber (see Fig.~\ref{fig:basic}).
An agent is spawned in the center of the room's longer wall. A stationary
monster is spawned at a random position along the~opposite wall.
The agent can strafe left and right, or shoot. A single hit is enough
to kill the~monster. The episode ends when the monster is eliminated
or after $300$ frames, whatever comes first. The agent scores $101$
points for killing the monster, $-5$ for a missing shot, and, additionally,
$-1$ for each action. The scores motivate the learning agent to eliminate
the monster as quickly as possible, preferably with a single shot\footnote{See also \href{https://youtu.be/fKHw3wmT_uA}{https://youtu.be/fKHw3wmT\_{}uA}}.

\subsubsection{Deep Q-Learning\label{subsec:Deep-Q-Learning}}

The learning procedure is similar to the Deep Q-Learning introduced
for Atari 2600 \cite{mnih-dqn-2015}. The problem is modeled as a
Markov Decision Process and Q-learning \cite{watkins:mlj92} is used
to learn the policy. The action is selected by an $\epsilon$-greedy
policy with linear $\epsilon$ decay. The Q-function is approximated
with a convolutional neural network, which is trained with Stochastic
Gradient Decent. We also used experience replay but no target network
freezing (see \cite{mnih-dqn-2015}). 

\begin{figure*}
\includegraphics[width=1\textwidth]{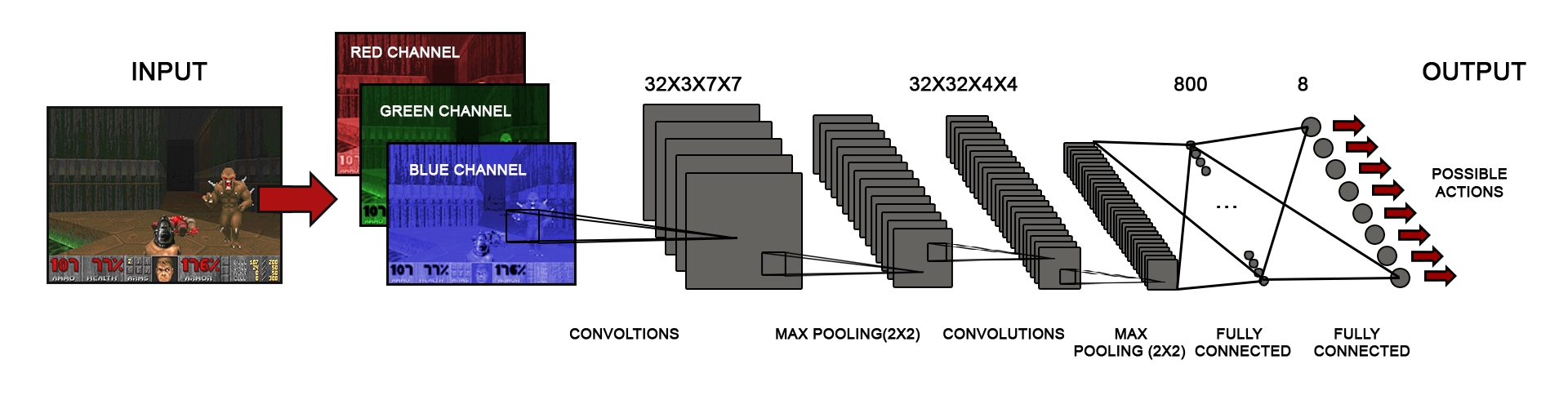} \caption{\label{fig:network}Architecture of the convolutional neural network
used for the experiment.}
\end{figure*}

\subsubsection{Experimental Setup}

\paragraph{Neural Network Architecture}

The network used in the experiment consists of two convolutional layers
with $32$ square filters, $7$ and $4$ pixels wide, respectively
(see Fig.~\ref{fig:network}). Each convolution layer is followed
by a max-pooling layer with max pooling of size $2$ and rectified
linear units for activation \cite{AISTATS2011_GlorotBB11}. Next,
there is a fully-connected layer with $800$ leaky rectified linear
units \cite{Maas2013} and an output layer with $8$ linear units
corresponding to the $8$ combinations of the $3$ available actions
(left, right and shot)%
.

\paragraph{Game Settings}

A state was represented by the most recent frame, which was a $60\times45$
$3$-channel RGB image. The number of skipped frames is controlled
by the \emph{skipcount} parameter. We experimented with skipcounts\emph{
}of $0$-$7$, $10$, $15$, $20$, $25$, $30$, $35$ and $40$.
It is important to note that the agent repeats the last decision on
the skipped frames.

\paragraph{Learning Settings}

We arbitrarily set the discount factor $\gamma=0.99$, learning rate
$\alpha=0.01$, replay memory capacity to $10\,000$ elements and
mini-batch size to $40$. The initial $\epsilon=1.0$ starts to decay
after $100\,000$ learning steps, finishing the decay at $\epsilon=0.1$
at $200\,000$ learning steps.

Every agent learned for $600\,000$ steps, each one consisting of
performing an action, observing a transition, and updating the network.
To monitor the learning progress, $1000$ testing episodes were played
after each $5000$ learning steps. Final controllers were evaluated
on $10\,000$ episodes. The experiment was performed on Intel Core
i7-4790k 4GHz with GeForce GTX 970, which handled the neural network.

\subsubsection{Results}

\begin{figure}
\begin{centering}
\includegraphics[width=1\columnwidth]{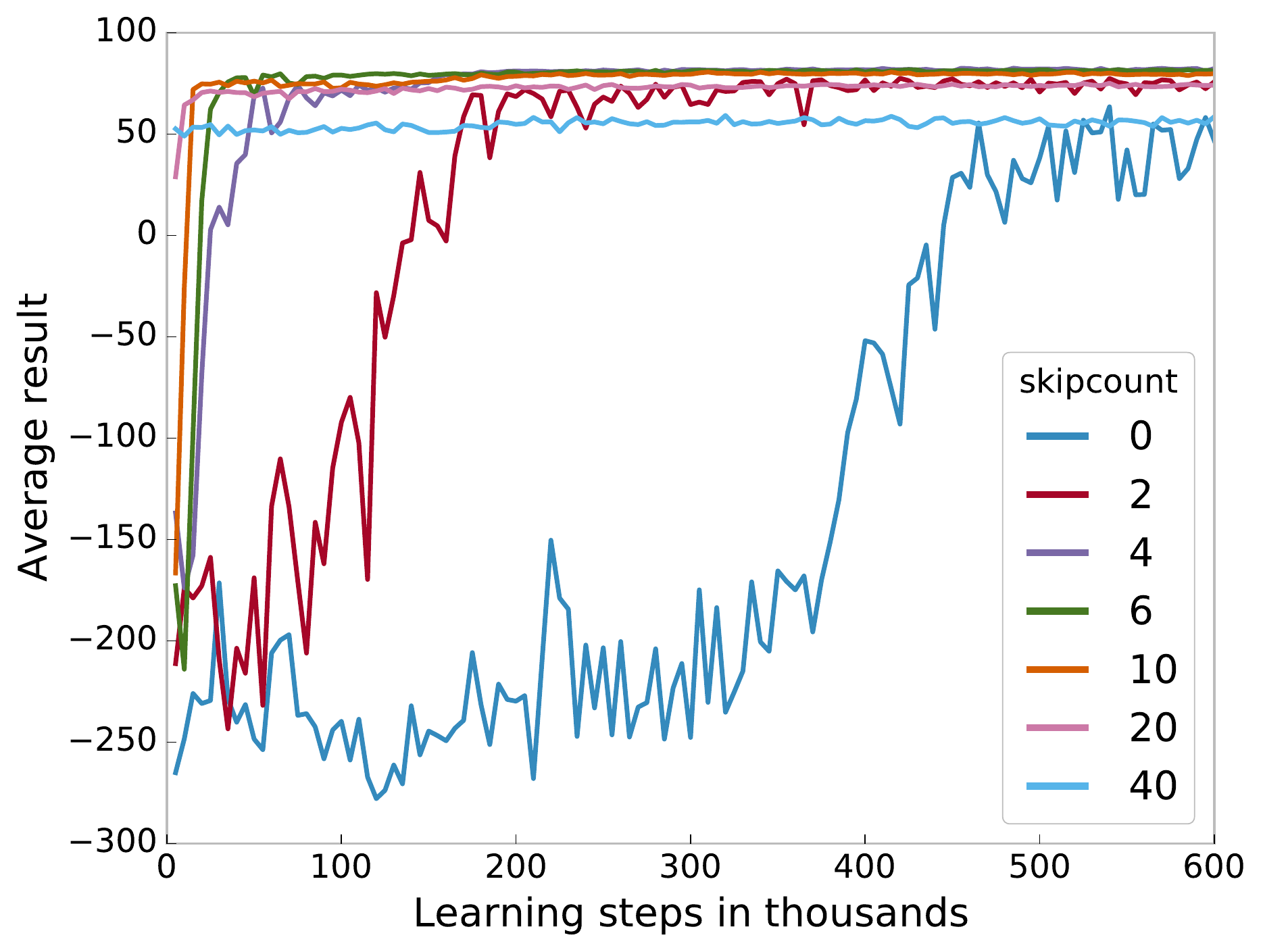} 
\par\end{centering}
\centering{}\caption{\label{fig:skiprate_time}Learning dynamics depending on the number
of skipped frames.}
\end{figure}

Figure \ref{fig:skiprate_time} shows the learning dynamics for the
selected skipcounts. It demonstrates that although all the agents
improve over time, the skips influence the learning speed, its smoothness,
as well as the final performance. When the agent does not skip any
frames, the learning is the slowest. Generally, the larger the skipcount,
the faster and smoother the learning is. We have also observed that
the agents learning with higher skipcounts were less prone to irrational
behaviors like staying idle or going the direction opposite to the
monster, which results in lower variance on the plots. On the other
hand, too large skipcounts make the agent `clumsy' due to the lack
of fine-grained control, which results in suboptimal final scores. 

The detailed results, shown in Table \ref{tab:SkiprateResults}, indicate
that the optimal skipcount for this scenario is $4$ (the ``native''
column). However, higher values (up to $10$) are close to this maximum. 

We have also checked how robust to skipcounts the agents are. For
this purpose, we evaluated them using skipcounts different from ones
they had been trained with. Most of the agents performed worse than
with their ``native'' skipcounts. The least robust were the agents
trained with skipcounts less than $4$. Larger skipcounts resulted
in more robust agents. Interestingly, for skipcounts greater than
or equal to $30$, the agents score better on skipcounts lower than
the native ones. Our best agent that was trained with skipcount $4$
was also the best when executed with skipcount $0$. 

It is also worth showing that increasing the skipcount influences
the total learning time only slightly. The learning takes longer primarily
due to the higher total overhead associated with episode restarts
since higher skipcounts result in a greater number of episodes.

To sum up, the skipcounts in the range of $4$-$10$ provide the best
balance between the learning speed and the final performance. The
results also indicate that it would be profitable to start learning
with high skipcounts to exploit the steepest learning curve and gradually
decrease it to fine-tune the performance.

\selectlanguage{english}%
\begin{table}
\begin{centering}
\caption{\label{tab:SkiprateResults}\foreignlanguage{american}{Agents' final
performance in the function of the number of skipped frames (`native').
All the agents were also tested for skipcounts$\in\{0,10\}$.}}
\par\end{centering}
\centering{}\scalebox{0.7}{
\def\arraystretch{1.2}
\sisetup{separate-uncertainty=true}
\begin{tabular}{
 S[table-format=2.0]
 S[table-format=2.1, table-figures-uncertainty=3, detect-weight]
 S[table-format=2.1, table-figures-uncertainty=3, detect-weight]
 S[table-format=2.1, table-figures-uncertainty=3, detect-weight]
 S[table-format=6.0, detect-weight]
 S[table-format=2.1, detect-weight]
}
\toprule
\multirow{2}{*}{skipcount} & \multicolumn{3}{c}{average score $\pm$ stdev} & \multirow{2}{*}{episodes} & \multirow{2}{*}{learning time [min]} \\
 & {native} & {0} & {10} &  & \\
\midrule
0 & 51.5\pm74.9 & 51.5\pm74.9 & 36\pm103.6 & 6961 & 91.1 \\
1 & 69\pm34.2 & 69.2\pm26.9 & 39.6\pm93.9 & 29378 & 93.1 \\
2 & 76.2\pm15.5 & 71.8\pm18.1 & 47.9\pm47.6 & 49308 & 91.5 \\
3 & 76.1\pm14.6 & 75.1\pm15 & 44.1\pm85.4 & 65871 & 93.4 \\
4 & \bfseries 82.2\pm9.4 & \bfseries 81.3\pm11 & 76.5\pm17.1 & 104796 & 93.9 \\
5 & 81.8\pm10.2 & 79\pm13.6 & 75.2\pm19.9 & 119217 & 92.5 \\
6 & 81.5\pm9.6 & 78.7\pm14.8 & 76.3\pm16.5 & 133952 & 92 \\
7 & 81.2\pm9.7 & 77.6\pm15.8 & 76.9\pm17.9 & 143833 & 95.2 \\
10 & 80.1\pm10.5 & 75\pm17.6 & \bfseries 80.1\pm10.5 & 171070 & 92.8 \\
15 & 74.6\pm14.5 & 71.2\pm16 & 73.5\pm19.2 & 185782 & 93.6 \\
20 & 74.2\pm15 & 73.3\pm14 & 71.4\pm20.7 & 240956 & 94.8 \\
25 & 73\pm17 & 73.6\pm15.5 & 71.4\pm20.8 & 272633 & 96.9 \\
30 & 61.4\pm31.9 & 69.7\pm19 & 68.9\pm24.2 & 265978 & 95.7 \\
35 & 60.2\pm32.2 & 69.5\pm16.6 & 65.7\pm26.1 & 299545 & 96.9 \\
40 & 56.2\pm39.7 & 68.4\pm19 & 68.2\pm22.8 & 308602 & 98.6 \\
\bottomrule
\end{tabular}
}
\end{table}

\selectlanguage{american}%

\subsection{Medikit Collecting Experiment}

The previous experiment was conducted on a simple scenario which was
closer to a 2D arcade game rather than a true 3D virtual world. That
is why we decided to test if similar deep reinforcement learning methods
would work in a more involved scenario requiring substantial spatial
reasoning.

\subsubsection{Scenario}

\begin{figure}
\begin{centering}
\includegraphics[width=1\columnwidth]{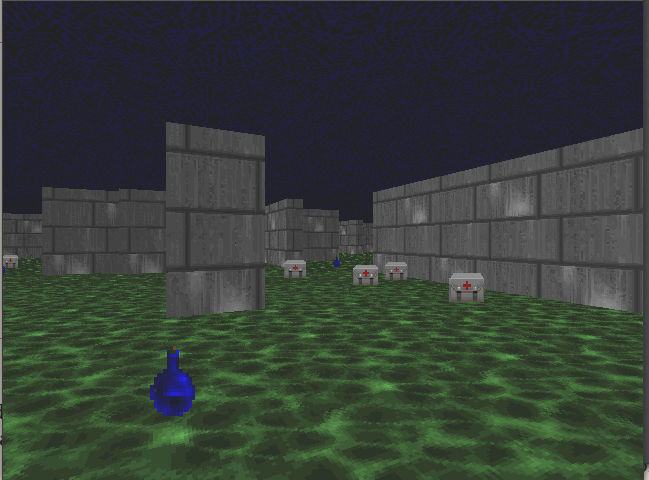}
\par\end{centering}
\caption{\label{fig:superhealth}Health gathering scenario}
\end{figure}

In this scenario, the agent is spawned in a random spot of a maze
with an acid surface, which slowly, but constantly, takes away the
agent's life (see Fig.~\ref{fig:superhealth}). To survive, the agent
needs to collect medikits and avoid blue vials with poison. Items
of both types appear in random places during the episode. The agent
is allowed to move (forward/backward), and turn (left/right). It scores
$1$ point for each tick, and it is punished by $-100$ points for
dying. Thus, it is motivated to survive as long as possible. To facilitate
learning, we also introduced shaping rewards of $100$ and $-100$
points for collecting a medikit and a vial, respectively. The shaping
rewards do not count to the final score but are used during the agent's
training helping it to `understand' its goal. Each episode ends after
$2100$ ticks ($1$ minute in real-time) or when the agent dies so
$2100$ is the maximum achievable score. Being idle results in scoring
$284$ points.

\subsubsection{Experimental Setup}

The learning procedure was the same as described in Section~\ref{subsec:Deep-Q-Learning}
with the difference that for updating the weights RMSProp \cite{Tieleman2012}
this time.

\paragraph{Neural Network Architecture}

  The employed network is similar the one used in the previous experiment.
The differences are as follows. It involves three convolutional layers
with  $32$ square filters $7$, $5$, and $3$ pixels wide, respectively.
The fully-connected layer uses $1024$ leaky rectified linear units
and the output layer $16$ linear units corresponding to each combination
of the $4$ available actions.

\paragraph{Game Settings}

The game's state was represented by a $120\times45$ $3$-channel
RGB image, health points and the current tick number (within the episode).
 Additionally, a kind of memory was implemented by making the agent
use $4$ last states as the neural network's input. The nonvisual
inputs (health, ammo) were fed directly to the first fully-connected
layer. Skipcount of $10$ was used.

\paragraph{Learning Settings}

We set the discount factor $\gamma=1$, learning rate $\alpha=0.00001$,
replay memory capacity to $10\,000$ elements and mini-batch size
to $64$. The initial $\epsilon=1.0$ started to decay after $4\,000$
learning steps, finishing the decay at $\epsilon=0.1$ at $104\,000$
episodes.

The agent was set to learn for $1000\,000$ steps. To monitor the
learning progress, $200$ testing episodes were played after each
$5000$ learning steps. The whole learning process, including the
testing episodes, lasted $29$ hours.

\subsubsection{Results}

\begin{figure}
\begin{centering}
\includegraphics[width=1\columnwidth]{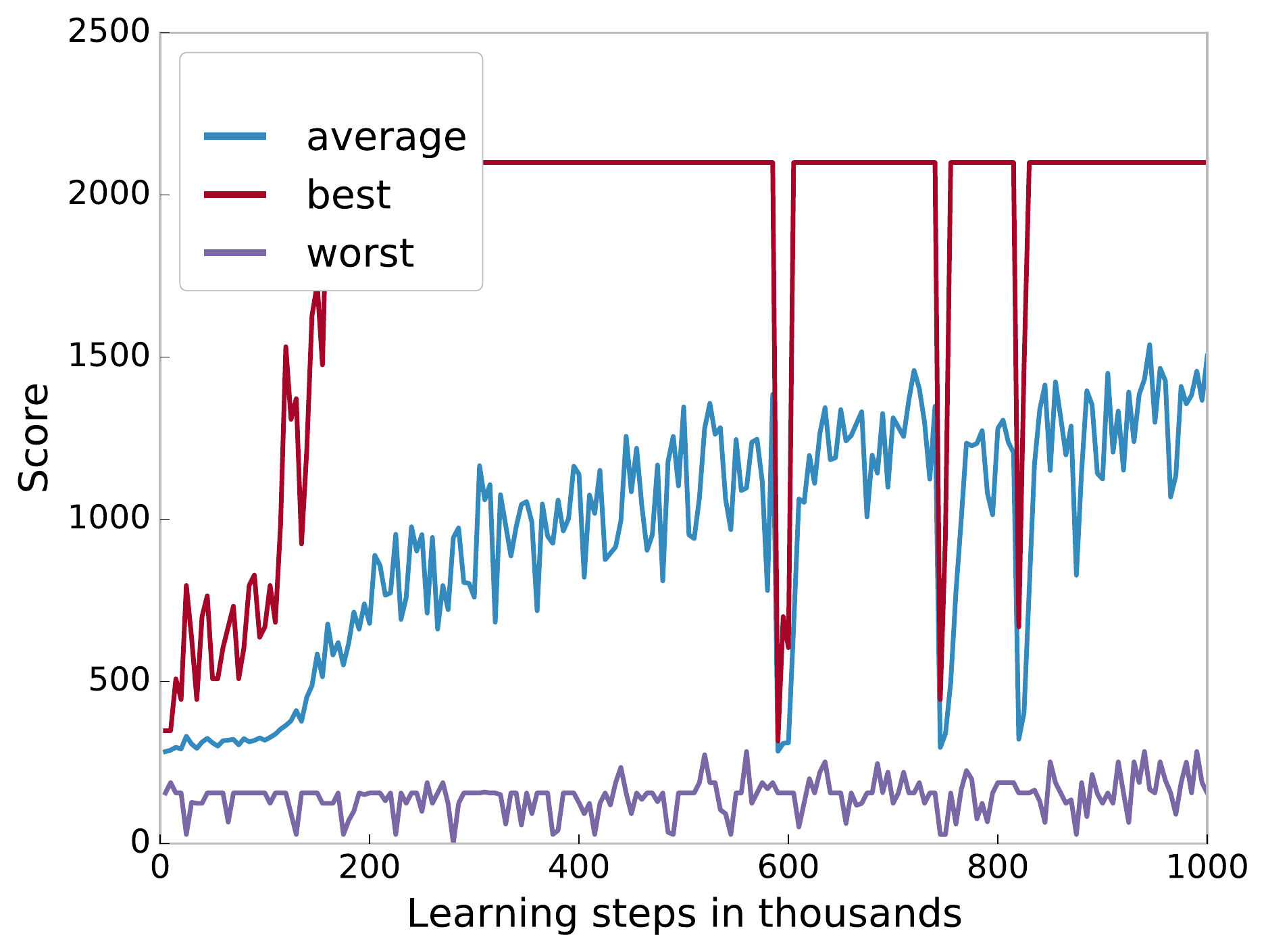}
\par\end{centering}
\caption{\label{fig:superhealth-dynamics}Learning dynamics for health gathering
scenario.}

\end{figure}

The learning dynamics is shown in Fig.~\ref{fig:superhealth-dynamics}.
It can be observed that the agents fairly quickly learns to get the
perfect score from time to time. Its average score, however, improves
slowly reaching $1300$ at the end of the learning. The trend might,
however, suggest that some improvement is still possible given more
training time. The plots suggest that even at the end of learning,
the agent for some initial states fails to live more than a random
player. 

It must, however, be noted that the scenario is not easy and even
from a human player, it requires a lot of focus. It is so because
the medikits are not abundant enough to allow the bots to waste much
time.

Watching the agent play\footnote{\href{https://www.youtube.com/watch?v=re6hkcTWVUY}{https://www.youtube.com/watch?v=re6hkcTWVUY}}
revealed that it had developed a policy consistent with our expectations.
It navigates towards medikits, actively, although not very deftly,
avoids the poison vials, and does not push against walls and corners.
It also backpedals after reaching a dead end or a poison vial. However,
it very often hesitates about choosing a direction, which results
in turning left and right alternately on the spot. This quirky behavior
is the most probable, direct cause of not fully satisfactory performance.

Interestingly, the learning dynamics consists of three sudden but
ephemeral drops in the average and best score. The reason for such
dynamics is unknown and it requires further research.

\section{Conclusions\label{sec:Conclusions}}

ViZDoom is a Doom-based platform for research in vision-based reinforcement
learning. It is easy-to-use, highly flexible, multi-platform, lightweight,
and efficient. In contrast to the other popular visual learning environments
such as Atari 2600, ViZDoom provides a 3D, semi-realistic, first-person
perspective virtual world. ViZDoom's API gives the user full control
of the environment. Multiple modes of operation facilitate experimentation
with different learning paradigms such as reinforcement learning,
apprenticeship learning, learning by demonstration, and, even the
`ordinary', supervised learning. The strength and versatility of environment
lie in is customizability via the mechanism of scenarios, which can
be conveniently programmed with open-source tools.

We also demonstrated that visual reinforcement learning is possible
in the 3D virtual environment of ViZDoom by performing experiments
with deep Q-learning on two scenarios. The results of the simple move-and-shoot
scenario, indicate that the speed of the learning system highly depends
on the number of frames the agent is allowed to skip during the learning.
We have found out that it is profitable to skip from $4$ to $10$
frames. We used this knowledge in the second, more involved, scenario,
in which the agent had to navigate through a hostile maze and collect
some items and avoid the others. Although the agent was not able to
find a perfect strategy, it learned to navigate the maze surprisingly
well exhibiting evidence of a human-like behavior.

ViZDoom has recently reached a stable $1.0.1$ version and has a potential
to be extended in many interesting directions. First, we would like
to implement a synchronous multiplayer mode, which would be convenient
for self-learning in multiplayer settings. Second, bots are now deaf
thus, we plan to allow bots to access the sound buffer. Lastly, interesting,
supervised learning experiments (e.g., segmentation) could be conducted
if ViZDoom automatically labeled objects in the scene.

\section*{Acknowledgment }

This work has been supported in part by the Polish National Science
Centre grant no. DEC-2013/09/D/ST6/03932. M. Kempka acknowledges the
support of Ministry of Science and Higher Education grant no. 09/91/DSPB/0602.

\bibliographystyle{plain}
\bibliography{FPS-research,bibliography,wjaskowski,all,library}

\end{document}